\documentclass{article}
\usepackage{ijcai23}

\usepackage{times}
\usepackage{soul}
\usepackage{url}
\usepackage[hidelinks]{hyperref}
\usepackage[utf8]{inputenc}
\usepackage[small]{caption}
\usepackage{graphicx}
\usepackage{amsmath}
\usepackage{amsthm}
\usepackage{booktabs}
\usepackage{algorithm}
\usepackage{algorithmic}
\usepackage[switch]{lineno}
\usepackage{graphicx}
\usepackage{hhline}

\title{Towards Equitable Representation in Text-to-Image Synthesis Models with the Cross-Cultural Understanding Benchmark (CCUB) Dataset}

\author{
    Zhixuan Liu \textsuperscript{\rm 1} \thanks{indicates equal contribution.},
    Youeun Shin \textsuperscript{\rm 1, \rm 2} \footnotemark[1],
    Beverley-Claire Okogwu \textsuperscript{\rm 1},
    Youngsik Yun \textsuperscript{\rm 1, \rm 2}, \\
    Lia Coleman \textsuperscript{\rm 1},
    Peter Schaldenbrand \textsuperscript{\rm 1}\thanks{indicates corresponding authors.},
    Jihie Kim \textsuperscript{\rm 2} \footnotemark[2],
    Jean Oh \textsuperscript{\rm 1} \footnotemark[2]
    \affiliations
    \textsuperscript{\rm 1} The Robotics Institute, Carnegie Mellon University\\
    \textsuperscript{\rm 2} Department of Artificial Intelligence, Dongguk University
    \emails
    \{zhixuan2, youeuns, bokogwu, youngsiy, liac, pschalde\}@andrew.cmu.edu,\\ jihie.kim@dgu.edu, jeanoh@cmu.edu
}

\begin{document}

\maketitle

\begin{abstract}
    It has been shown that accurate representation in media improves the well-being of the people who consume it. By contrast, inaccurate representations can negatively affect viewers and lead to harmful perceptions of other cultures. 
    To achieve inclusive representation in generated images, 
    we propose a culturally-aware priming approach for text-to-image synthesis using a small but culturally curated dataset that we collected, known here as Cross-Cultural Understanding Benchmark (CCUB) Dataset, to fight the bias prevalent in giant datasets. 
    Our proposed approach is
    comprised of two fine-tuning techniques: (1) Adding visual context via fine-tuning a pre-trained text-to-image synthesis model, Stable Diffusion, on the CCUB text-image pairs, and 
    (2) Adding semantic context via automated prompt engineering using the fine-tuned large language model, GPT-3, trained on our CCUB culturally-aware text data. 
    CCUB dataset is curated and our approach is evaluated by people who have a personal relationship with that particular culture. Our experiments indicate that priming using both text and image is effective in improving the cultural relevance and decreasing the offensiveness of generated images while maintaining quality. 
    Our CCUB dataset and codes\footnote{\url{https://github.com/cmubig/CCUB}} are publicly available.
\end{abstract}

\section{Introduction}

\begin{figure}[t!]
\centering
\includegraphics[width=1\columnwidth]{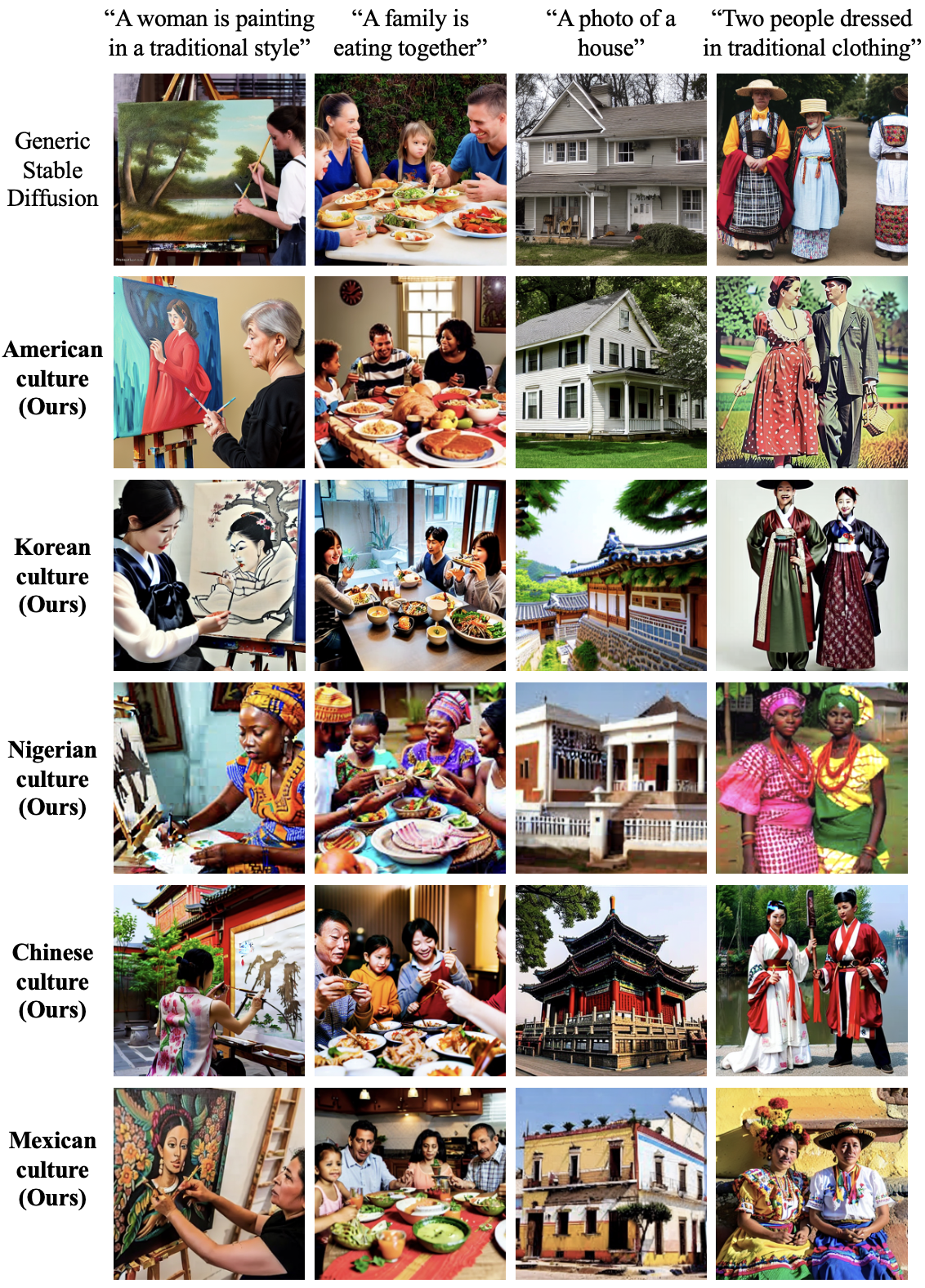} 
\caption{Sample images generated for five different countries by our proposed culturally-aware text-to-image synthesis approach; the images in the first row show the results from the generic Stable Diffusion as references.
}%
\label{fig::baseline}%
\end{figure}

Representation matters. In media, studies repeatedly show that representation affects the well-being of its viewers~\cite{shaw2010representationInVideoGames,caswell2017representation,elbaba2019teenMediaRepr}.  
Representation can positively affect viewers by providing them with role models that they identify with, but it can also negatively affect viewers by creating harmful, stereotypical understandings of people and culture~\cite{castaneda2018powerRepresentation}.
When people are accurately represented in media, it allows people to properly understand cultures without harmful stereotypes forming~\cite{dixon2000overrepresentation,mastro2000portrayal}.
Despite the benefits of representation, many media generating Artificial Intelligence (AI) models show poor representation in their results~\cite{ntoutsi2020bias}.  Many of these issues stem from their large training datasets which are gathered by crawling the Internet without filtering supervision and contain malign stereotypes and ethnic slurs among other problematic content~\cite{birhane2021stereotypesInLAION}. 

As AI models are increasingly used to create and aid in the production of visual content, it is important that the models have a true understanding of culture such that it can give accurate and proper representation leading to well-being rewards for its consumers.
In this paper, we aim to address such a representation issue in image generation and introduce a new task of \textit{culturally-aware} image synthesis: generating visual content within a cultural context that is both accurate and inoffensive. Our overarching goal is to improve the well-being of consumers of the AI generated images with particular attention to those consumers from underrepresented groups. Specifically, we formulate the culturally-aware text-to-image synthesis task to take an additional input of a country name to specify a cultural context in addition to language description. 

It was found that large datasets such as the LAION-5B~\cite{schuhmann2021laion} used to train many text-to-image synthesis models such as Stable Diffusion~\cite{rombach2021stableDiffusionOriginal} are Anglo-centric and Euro-centric~\cite{birhane2021stereotypesInLAION} 
as shown in the top row of Figure~\ref{fig::baseline}. 
As a consequence, these powerful models may generate culturally offensive images due to misrepresentation during training. 
Our research question is, how can effective existing text-to-image models be improved to become more culturally representative and thus less offensive? It may be infeasible to vet billions of training examples for accurate cultural content.

We hypothesize that a small dataset that is veritably representative of a culture can be used to prime pre-trained text-to-image models to guide the model towards more culturally accurate content creation.  To verify the hypothesis, we collected a dataset of image and caption pairs for 8 cultures.  For each culture, data was collected by a few people who are native of that culture 
as they are the people who properly understand it and are most affected by its misrepresentations. We call this the Cross-Cultural Understanding Benchmark (CCUB) dataset which comprises of 100-200 images each with a manually written caption as shown in Figure~\ref{fig::dataset}.


We propose two techniques for enhancing the text-to-image pipelines using  CCUB.
First, we fine-tune a text-to-image synthesis model, Stable Diffusion,  on the CCUB text-image pairs to generate images tailored for a given cultural context.
Second, we create an automatic prompt augmenting approach using GPT-3~\cite{Brown2020LanguageMA} fine-tuned on CCUB to include culturally relevant details, e.g., 
``Two people walking down a street'' can be augmented with ``\textit{using WeChat Pay to pay a bus ticket, in Shenzhen, China}."

We evaluate our approach's two components individually as well as combined against the baseline of simply specifying the culture in the text prompt.  Our evaluation was performed by native people of each country. 
Our survey results based on 2,244 image comparisions conducted by 72 participants from 5 countries indicate that our proposed approach is both less offensive and more cultural relevant than simply adding the country name as a suffix to the prompt. 
Our contributions are as follows:
\begin{enumerate}
    \item The introduction of culturally-aware text-to-image synthesis as a valuable task within text-to-image synthesis;
    \item The Cross-Cultural Understanding Benchmark (CCUB) dataset consisting of 1,095 culturally representative image-text pairs across 8 countries; and 
    \item Two techniques for culturally customizing a text-to-image synthesis model.
\end{enumerate}

\section{Related Work}
\subsection{Cultural Datasets}
Various efforts have been made to build a dataset that contains a precise representation of each culture around the world, especially for the underrepresented groups and smaller populations, to combat the bias of benchmark datasets. MaRVL dataset~\cite{Liu2021VisuallyGR} created a set of cultures and languages, including Indonesian, Swahili, Tamil, Turkish, and Mandarin Chinese, comprised of diverse cultural concepts to mitigate existing North American or Western European bias. While sharing the similar intuition, MaRVL was specifically developed for the reasoning task covering common, popular concepts only. 

Dollar Street~\cite{Rojas2022TheDS} aimed to capture accurate demographic information based on socioeconomic features, such as everyday household items and monthly income, of 63 countries worldwide. 
However, this dataset gives less diverse scenarios. Most of the images in this dataset provide indoor views with limited cultural features. 


\subsection{Culturally Conditioned Machine Learning}

Accurately representing culture with Machine Learning is an open challenge.
Many models, such as Craiyon~\cite{dayma2021dallMiniCraiyon} fail to capture certain distinguishing features relating to a country's dominant culture~\cite{reviriego2022text}. One method to address this involves the inclusion of semantic understanding in a model such as the ERNIE-ViLG 2.0.~\cite{feng2022ernie}. 
A similar approach can be seen in Japanese Stable Diffusion~\cite{japanse2022stableDiffusion}, which fine tunes the Stable Diffusion U-Net and retrains the text encoder on the 100 million images with Japanese caption within the LAION-5B~\cite{schuhmann2021laion} dataset.

While these approaches produce better cultural representations of Japan and China, it is not easy to be used universally. Adapting these approaches requires millions of training examples which cannot be easily met for cultures with less internet presence. Also, these  datasets are so large that it is infeasible to vet them for harmful and stereotypical information. Our approach strives for cultural representation using a dataset that is smaller (100-200 images) and hand selectable. 

\subsection{Modifying Text-to-Image Diffusion Models}
Diffusion-based text-to-image synthesis models have improved incredibly over the past year in image quality and language understanding; however, these models are still Anglo-centric and contain gender and racial biases at least in part due to the lack of supervision in their large text-image training datasets~\cite{birhane2021stereotypesInLAION}.
One way to address this problem while leveraging the knowledge obtained from the large dataset is to fine-tune the latent diffusion models. ~\cite{ruiz2022dreambooth} and ~\cite{gal2022imageWorthOneWord} 
propose textural inversion methods that allow the latent diffusion models to generate images with specific visual concepts. However, these methods restrict the generation to be an object or a style and cannot generalize on the expression of abstract concepts, for example, a culture. Inspired by the approaches of ~\cite{chambon2022adapting} and ~\cite{TextToPokemonGenerator}, the text-to-image diffusion model can generate domain-specific images by fine-tuning the U-Net of the model using a batch of data from that domain. 

\subsection{Prompt Engineering}

Originating from the field of Natural Language Processing (NLP), prompt engineering, which can be conceived as programming in natural language~\cite{promptprogrammingLLMs}, is to design the input text prompt to retrieve user-desired outcomes from language models. 
Inspired by the findings in the NLP domain, researchers in computer vision have been exploring the effects of prompt engineering. In order to present design guidelines for better outcomes in text-to-image generation models~\cite{Liu2022DesignGF}, several permutations of prompt engineering using a template were conducted in terms of subject and style in art. To discover some tricks and keywords to boost the quality of the output image in the image generation models such as DALL-E 2~\cite{Ramesh2022HierarchicalTI} and Midjourney~\cite{Midjourney}, various experiments through trial-and-error are ongoing on the Internet as well. Online community~\cite{PromptEngineeringFromWordstoArt} has come up with a prompt engineering template for artwork, consisting of terms regarding styles, artists, vibes, and perspectives.

\section{Cross-Cultural Understanding Benchmark (CCUB) Dataset}

\begin{table}[]
\centering
\resizebox{\columnwidth}{!}{%
\begin{tabular}{|l||lllllllll||l|}
\hline
 &
  \rotatebox{90}{food \& drink} &
  \rotatebox{90}{clothing} &
  \rotatebox{90}{artwork} &
  \rotatebox{90}{dance \& music} &
  \rotatebox{90}{religion} &
  \rotatebox{90}{architecture} &
  \rotatebox{90}{people} &
  \rotatebox{90}{city} &
  \rotatebox{90}{nature} &
  \rotatebox{90}{\textbf{total}}\\ \hline 
Korea & 55 & 9 & 10 & 18 & 3 & 23 & 12 & 22 & 7 & 159 \\ \hline 
Japan & 20 & 7 & 13 & 15 & 3 & 11 & 26 & 13 & 14 & 122 \\ \hline
China & 25 & 18 & 6 & 12 & 7 & 13 & 20 & 23 & 10 & 134 \\ \hline
Mexico & 22 & 14 & 23 & 10 & 6 & 18 & 19 & 15 & 7 & 134 \\ \hline
Nigeria & 16 & 19 & 12 & 11 & 8 & 19 & 12 & 31 & 5 & 133 \\ \hline
Norway & 24 & 11 & 7 & 14 & 5 & 20 & 18 & 20 & 13 & 132 \\ \hline
Vietnam & 27 & 14 & 10 & 15 & 7 & 14 & 17 & 16 & 10 & 130 \\ \hline
\begin{tabular}[c]{@{}l@{}}United\\  States\end{tabular} & 23 & 10 & 15 & 12 &7 & 17 & 22 & 29 & 16 & 151 \\ \hhline{|=||=========||=|}
\textbf{Total} & 212 & 102 & 96 & 107 & 46 & 135 & 146 & 169 & 82 & \textbf{1095} \\ \hline

\end{tabular}%
}
\caption{
This table shows the scale of our CCUB dataset. The number of hand-selected images and their corresponding captions in nine cultural categories for eight different cultures are listed.}
\label{tab:cultural-datasets}
\end{table}

\begin{figure}[t]
\centering
\includegraphics[width=1\columnwidth]{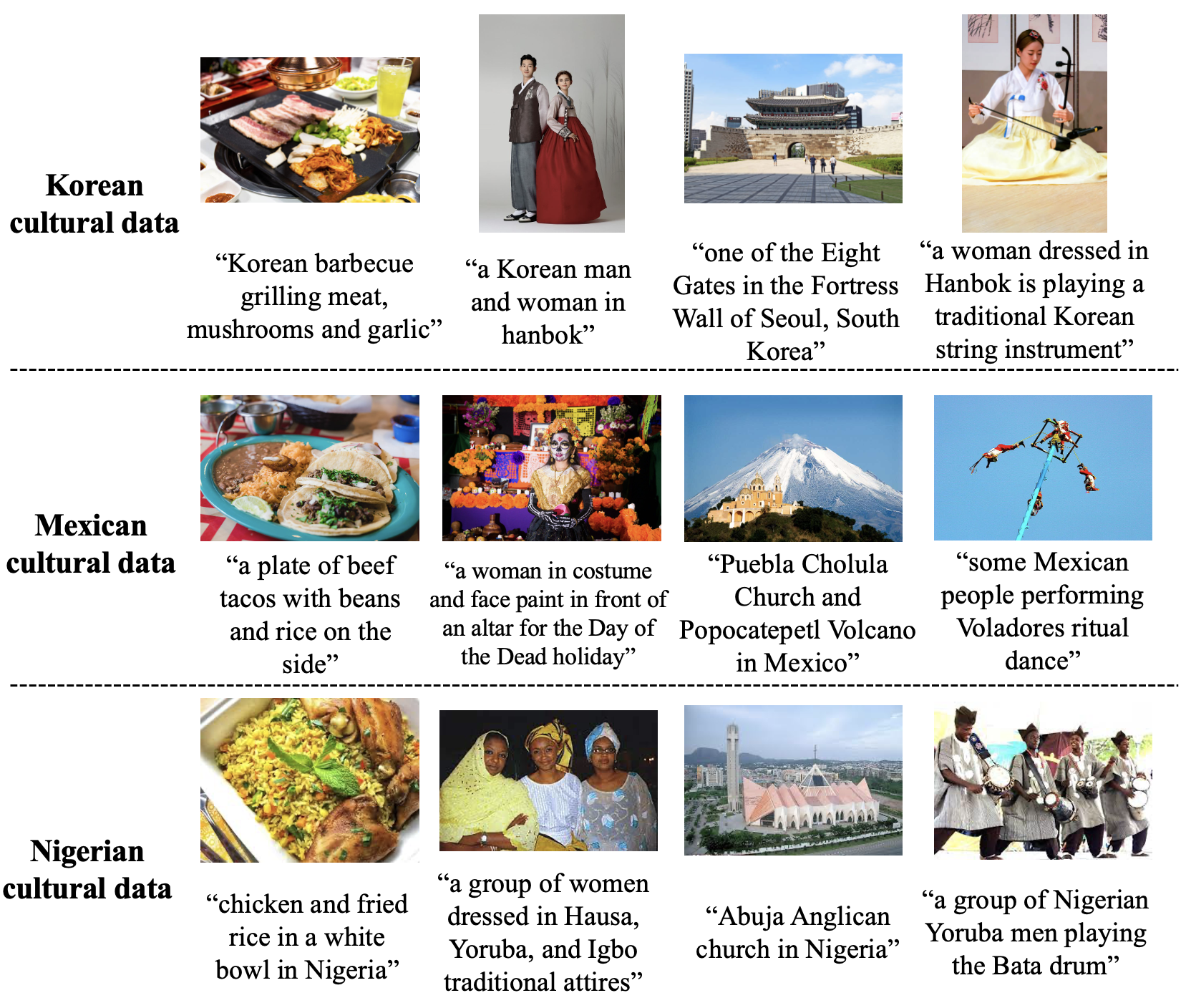} 
\caption{
Images and captions in our CCUB dataset for three of the eight countries whose cultures are represented in CCUB.
}
\label{fig::dataset}
\end{figure}

Our CCUB dataset provides text-image pairwise data across 8 cultures and is designed for text-to-image synthesis tasks. For each culture subset, the proposed dataset contains images among various scenarios. Moreover, the text data concisely describes the culturally aware content of each image thus giving the language model a clear guide towards cultural awareness.

\subsection{Culturally Representative Data Collection }
Following the definition of culture in~\cite{10.2307/2378980} and~\cite{ids}, nine categories are used to represent cultural elements in our dataset:  
food \& drink, clothing, artwork, dance and music, religion, architecture, people, city and nature. The categories are further divided into traditional and modern to reflect a characteristic of the culture that culture changes over time.

Our CCUB image  are collected based on the nine cultural categories. For collection, we recruited cultural experts who confidently know this culture well or belong to it. Cultural experts are asked to collect 10-20 relevant images containing different objects for each cultural category. The images were collected either from Creative Commons licensed images from Google searches or the collectors own photographs. Cultural experts were also asked to select images with common or culturally representative items. 

Each image in the CCUB dataset is also captioned by cultural experts forming paired image-text data.
Cultural experts were asked to focus on the general and specific items in each cultural image, rather than adding captions to subtle components of the image.
The captions accurately express cultural contents in English as opposed to large datasets such as LAION~\cite{schuhmann2021laion} which are scraped from the internet and not vetted for cultural accuracy.


\subsection{Properties}
Our CCUB dataset contains culturally representative image-text pairs in eight different cultures. 
As shown in Table \ref{tab:cultural-datasets}, our cultural dataset is in a minute scale considering that generic Stable Diffusion was trained on LAION-2B-EN that includes more than 2.3 billion text-image pairs.
Our table 
Figure \ref{fig::dataset} shows some selected samples of our CCUB dataset.


\section{Culturally-aware Text-to-image Synthesis}

\begin{figure*}
    \centering
    \includegraphics[width=\textwidth]{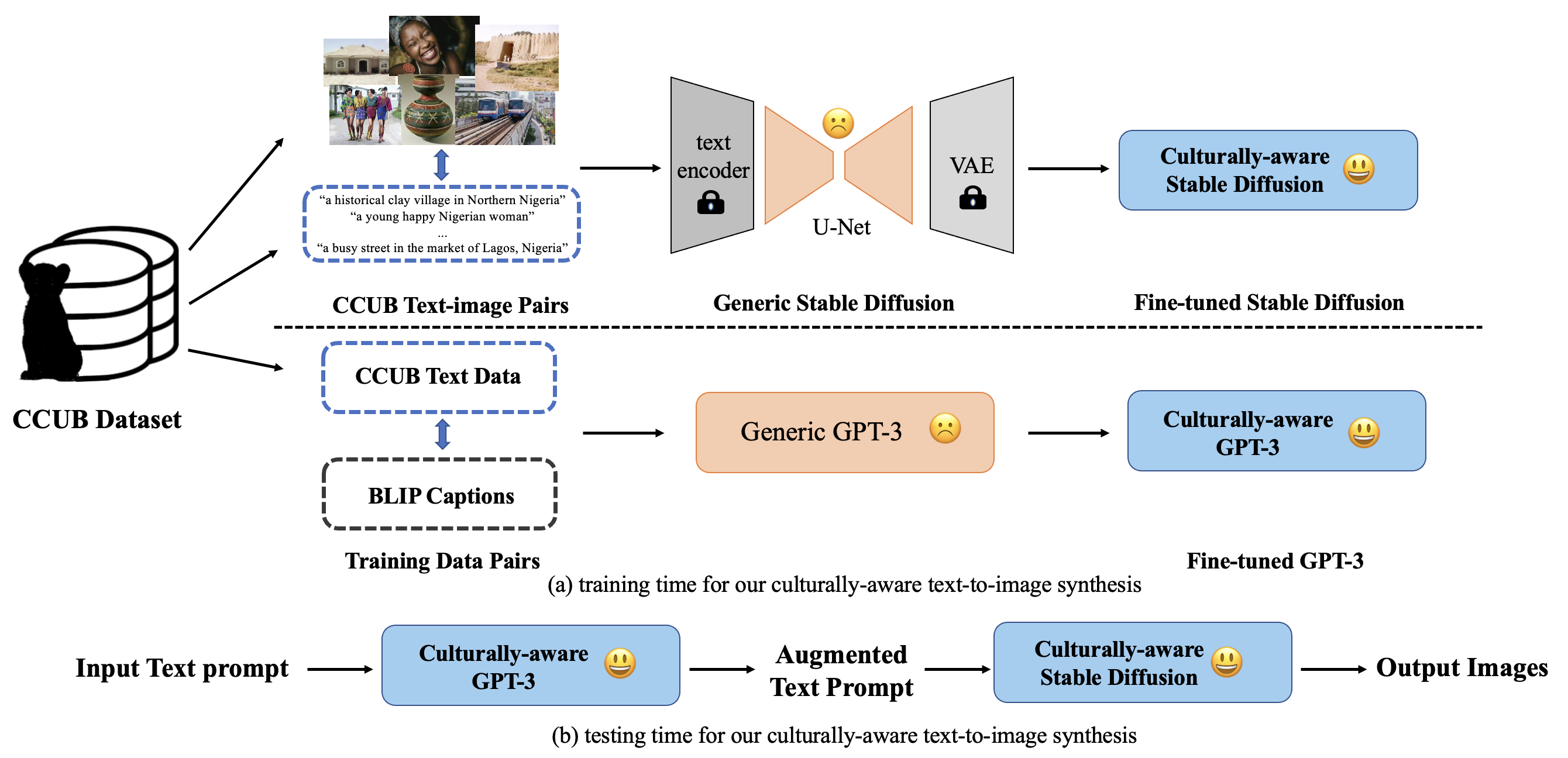}
    \caption{
    Figure (a) is the overview of our two fine-tuning techniques based on our CCUB dataset. Models with cultural biases are marked by sad faces, while culturally-aware models are marked by smiling faces. Figure (b) is the workflow for our approach toward culturally-aware text-to-image synthesis. 
    }
    \label{fig:approach}
\end{figure*}
\subsection{Fine-tuning Stable Diffusion}

The Stable Diffusion~\cite{rombach2021stableDiffusionOriginal} pipeline generates natural images under the condition of text prompts.  The input text prompt is firstly encoded using the CLIP~\cite{radford2021-clip} text encoder. A U-Net architecture model creates the output image encoding by denoising from random noise conditioned upon the encoded text. A Variational Autoencoder (VAE) converts the image encoding into a high-resolution image.

We alter Stable Diffusion to have a more accurate understanding of a given culture to address its known bias towards generating Western-focused imagery.
In our approach, following a similar approach to \cite{ruiz2022dreambooth} and ~\cite{chambon2022adapting}, the U-Net of the Stable Diffusion model is further trained on our CCUB image-text pairs
while keeping the text encoder and autoencoder (VAE) frozen. The fine-tuning of U-Net is equivalent to the training process of original Latent Diffusion Models (LDMs): by minimizing the LDM loss in several denoising time steps. The LDM loss is given by:

$$
L_{L D M}:=E_{z \sim \mathcal{E}(x), y, \epsilon \sim \mathcal{N}(0,1), t}\left[\left\|\epsilon-\epsilon_\theta\left(z_t, t, c\right)\right\|_2^2\right]
$$

where $t$ is the time step; $z_t$, the latent noise at time $t$; $c$, the text encoding of a text prompt; $\epsilon$, the noise sample; and $\epsilon_\theta$, the noise estimating U-Net model.

\textbf{Implementation detail}: We set the learning rate to be small (e.g., 1e-5) to slightly change parameters of the U-Net, and we run 150 epochs on the culturally representative CCUB image-text data pairs to fine-tune one Stable Diffusion. This generally leads to convergence of the images produced by our fine-tuned Stable Diffusion toward our culturally-aware training data.

\subsection{Prompt Augmenting}
To enhance the richness of cultural representation in the text domain,  we propose an approach to automatically augment the given text prompt with a fine-tuned large language model, trained further based on our culturally representative text data. Whereas current prompt augmentation in text-to-image synthesis is processed in a fixed set of manual templates, we benefit from the prior knowledge and generative ability of GPT-3~\cite{Brown2020LanguageMA} to describe the components of a specific culture in a more detailed way.

First, we prepare the training dataset for fine-tuning GPT-3 with pairs of a base prompt without culturally specific words and its corresponding culturally-aware prompt. We use BLIP~\cite{Li2022BLIPBL}, a unified vision-language understanding and generation model, to automatically caption our CCUB images. The four most relevant captions are generated per image, two based on beam search and the others by nucleus sampling, respectively. We then choose the best matching caption that shows the highest CLIP~\cite{radford2021-clip} score. The BLIP captioning had some errors, such as cultural misinterpretation, meaningless and repetitive words, or miscounted numbers. Such flaws are manually corrected before being used as the base prompt in pairs with our CCUB text captions.

We then fine-tune the largest GPT-3, Davinci, to translate between the BLIP captions and the CCUB captions. During fine-tuning, the model learns the latent concept of the specific cultural content and text format that needs to be generated. The base prompt is concatenated with ``\verb|\n\n###\n\n|'' as a separator token, and ``\verb|\nEND|'' is appended to complete the text as the stop token. During inference, we sample culturally augmented prompts given new input prompts using \verb|temperature=0.7|. Examples of augmented prompts are underlined in the row of prompt augmenting in Figure~\ref{fig:ablation1}.

\section{Experiments}\label{sec:experiments}
We produced surveys to evaluate the effectiveness of our two proposed techniques for culturally-aware text-to-image synthesis and compare them to a baseline of simply appending the culture to the prompt, e.g., ``A family eating dinner \textit{, China}." and using an existing text-to-image model.

In setting up our study, we consider a comparative structure between images: the baseline image versus another image from our results. 
The setup of a single question in our survey was as follows: Given two images, the participant selects which image best fits three given comparative properties. The properties analyzed were: (1)
\textbf{Text and Image Alignment}: Participants are given a text prompt and consider which of the two images is more similar to the prompt; 
(2)
\textbf{Cultural Alignment}: Participants decide which of the two images is a better representation of the country's culture; and 
(3)
\textbf{Offensiveness}: 
Participants consider which of the two images is more offensive to them. 

The participants for the study were selected based on whether they had a personal understanding of the culture for which the images in the survey were generated.
Participants were recruited among university students, friends, and family members of the authors.  It was ensured that the participants would not be able to discern the approaches used to generate the compared images by randomizing the order of questions and images in the survey.

\section{Results}

\subsection{Baselines}

\begin{figure}[t]
\centering
\includegraphics[width=\columnwidth]{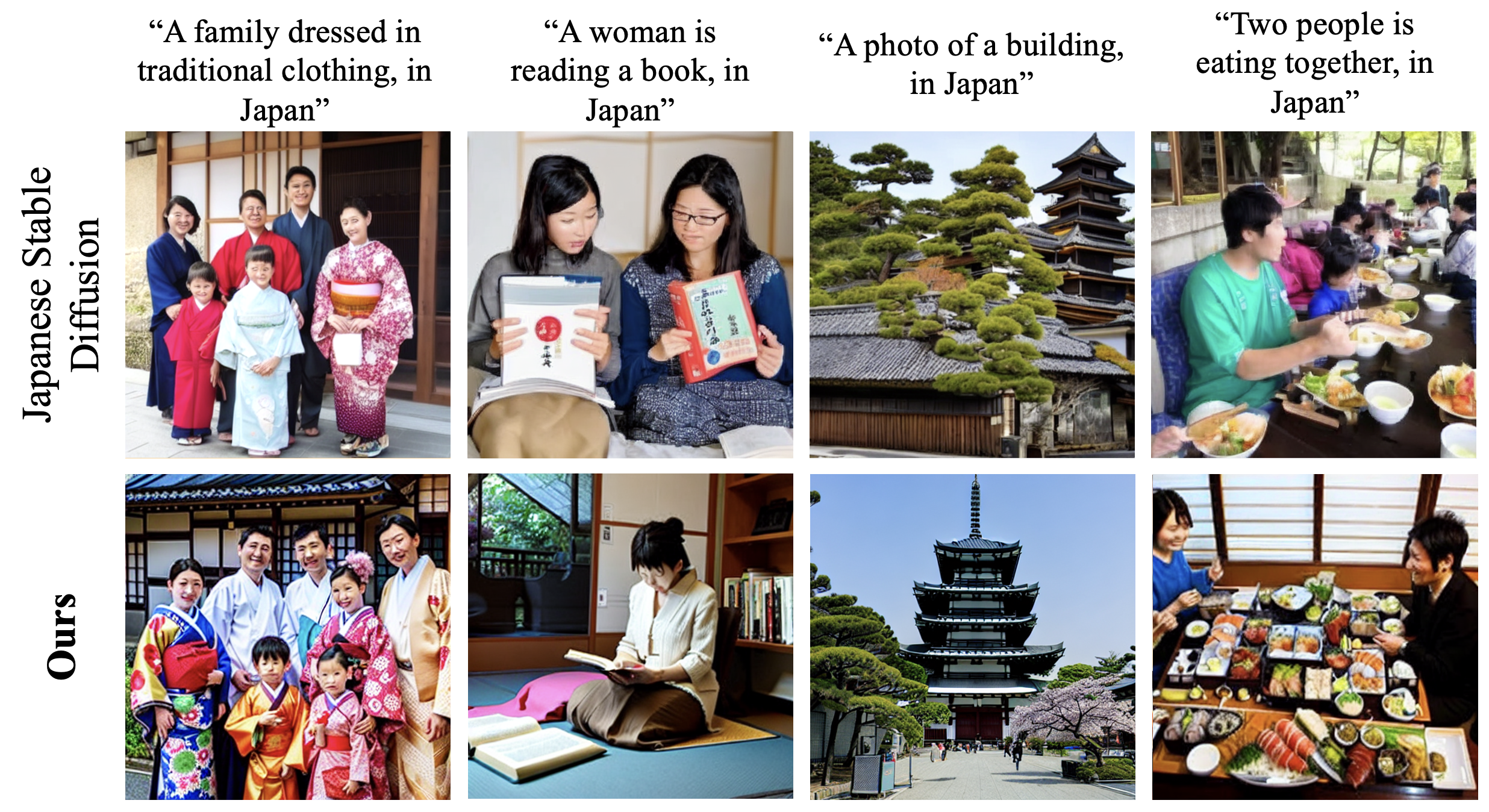} 
\caption{Compare between our (fine-tuned Stable Diffusion with prompt augmenting) method with Japanese Stable Diffusion. Top row shows the
language input used to generate images.}
\label{fig::Japanese_stable}
\end{figure}

\begin{figure}[t]
\centering
\includegraphics[width=1\columnwidth]{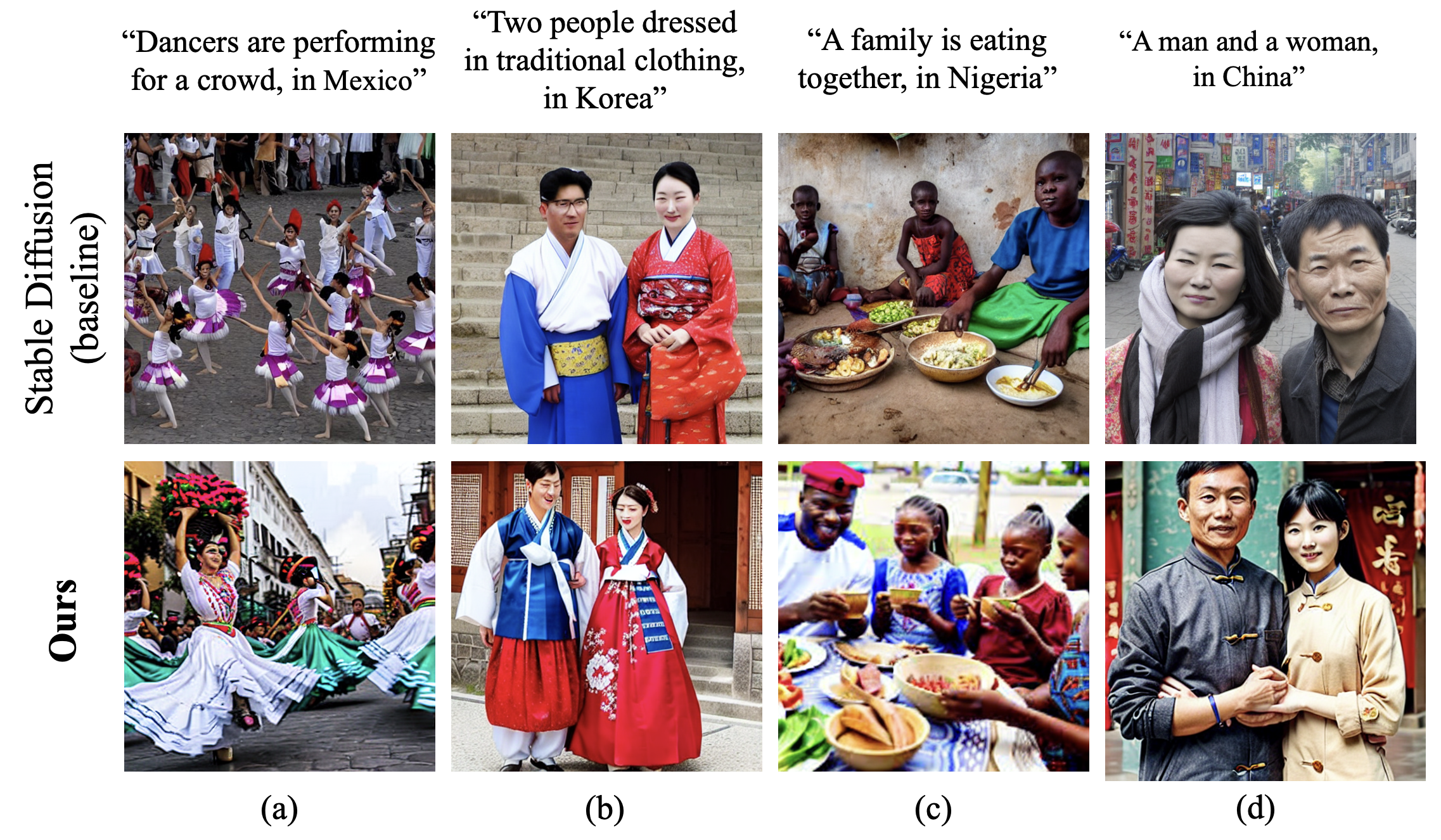} 
\caption{Comparison between Stable Diffusion and our approach with prompt augmentation and fine-tuning of Stable Diffusion method. The top row shows the
language input used to generate images.}
\label{fig::biases}
\end{figure}

\begin{figure*}
    \centering
    \includegraphics[width=14cm]{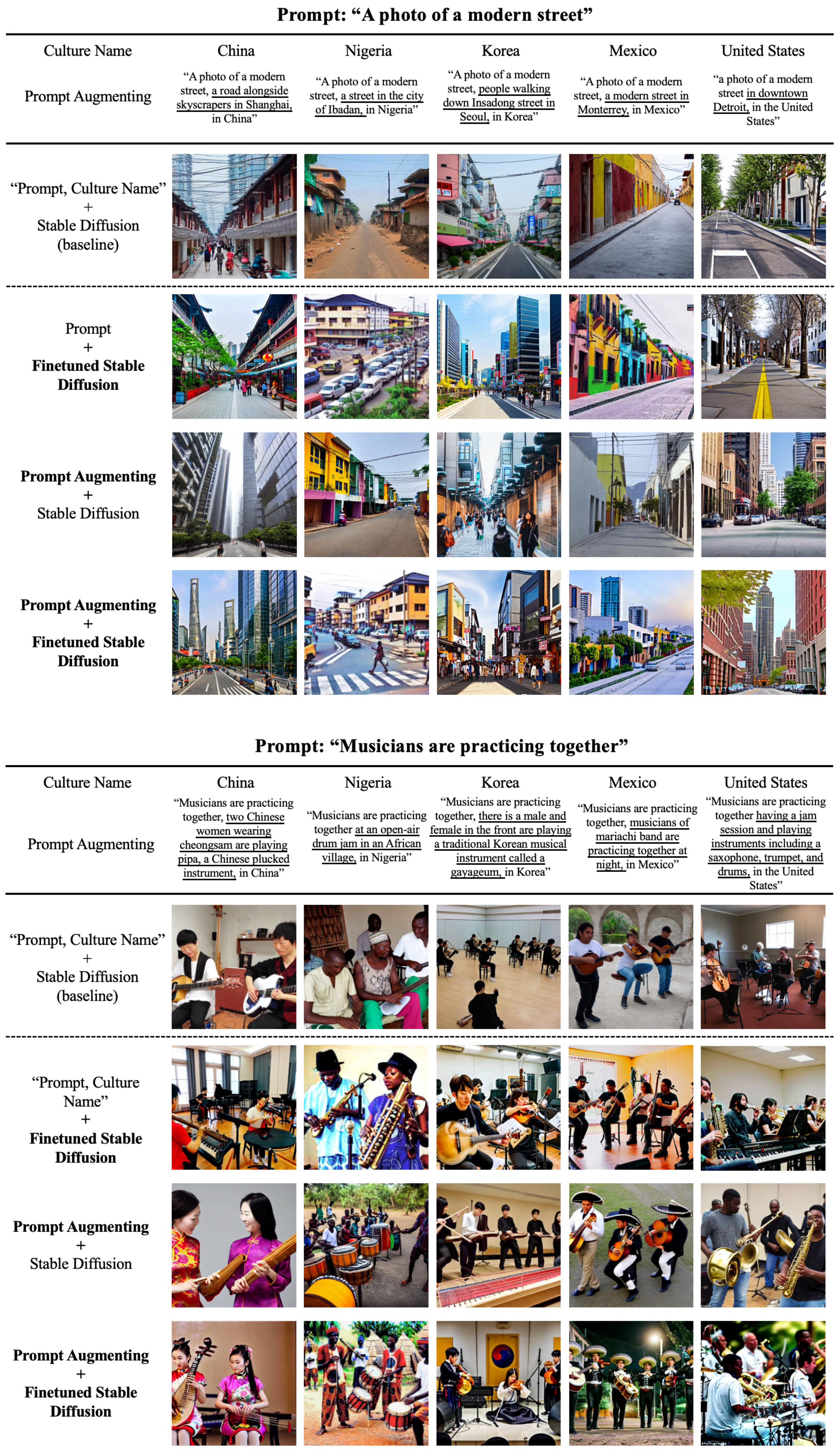}
    
    \caption{An ablation study of our proposed approach to culturally-aware text-to-image synthesis. The augmented prompts are underlined.
    }
    \label{fig:ablation1}
\end{figure*}

\noindent \textbf{Japanese Stable Diffusion:}
We compare our approach with Japanese Stable Diffusion~\cite{japanse2022stableDiffusion} in Figure \ref{fig::Japanese_stable}. 
Both approaches achieve similar results in terms of image-text alignment and culturally relevant information such as clothing, architecture, and food. While both approaches are qualitatively similar, the Japanese Stable Diffusion model was trained on 100 million Japanese image-text pairs for training while ours required only 100-200.

\noindent \textbf{Stable Diffusion:}
As a baseline, we used Stable Diffusion by simply adding a culture name as a suffix to the text prompt. The results in Figure \ref{fig::biases} show that this approach fails to capture cultural identity and instead produces many culturally biased images. We qualitatively report the following three issues:
(1) The baseline method generates images with elements that are still in the Western form as in Figure \ref{fig::biases} (a),  where the baseline method generated dancers performing Western ballet dance when the desired cultural context was Mexican,
while ours is able to generate images with more representative Mexican elements.
(2) The baseline method exhibits cultural misunderstanding, as shown in Figure \ref{fig::biases} (b), where images of Japanese style clothing was generated when the desired cultural context was Korean.
(3) The baseline generates images that incorporate cultural stereotypes and discrimination, as in Figure \ref{fig::biases} (c), the image generated by the baseline approach contains cultural biases and discrimination, failing to represent modern Nigerian culture. 
Also, in column (d), the baseline-generated image contains offensive, stereotypical elements of Chinese culture, especially in the facial features of the characters. 
All three of these issues are mitigated in our culturally-aware text-to-image synthesis approach.

\begin{table}[]
\centering
\small
\begin{tabular}{lccc}
              & \begin{tabular}[c]{@{}c@{}}Text-Image\\ Alignment $\uparrow$\end{tabular} & 
              \begin{tabular}[c]{@{}c@{}}Cultural-\\ Alignment $\uparrow$\end{tabular} &
              \begin{tabular}[c]{@{}c@{}}\\ Offensiveness $\downarrow$\end{tabular}  \\
Fine-Tuned SD & \textbf{66}                                                       & 67 & \textbf{33}                                                           \\
Prompt Aug.   & 43                                                                       & 57 & 50                                                              \\
Combined      & 57                                                                      & \textbf{71}  & 36                                                   
\end{tabular}
\caption{Percentage preference between the two proposed techniques (individual and combined) versus the baseline in our survey (Section ~\ref{sec:experiments}).}
\label{tab:quant_results}
\end{table}

\subsection{Qualitative Results}

Qualitative results from our two approaches versus a baseline are in Figure~\ref{fig:ablation1}. The augmented prompts are displayed at the top of each example and add culturally specific information to the prompts such as the city of Ibadan to a Nigerian prompt. These augmented prompts can generally help guide Stable Diffusion to generate something more culturally relevant given a text prompt without specific cultural information.

In general, faces in the fined tuned Stable Diffusion model look more natural as seen in the musicians examples of Figure~\ref{fig:ablation1}.  Many of the Stable Diffusion examples appear older photographs, and the fine-tuning  helps give the images a more contemporary appearance.  Fine-tuning also provides Stable Diffusion with additional image synthesis content capabilities, e.g., the baseline model was unable to produce the Chinese pipa instrument without the fine-tuning as seen in the lower left of Figure ~\ref{fig:ablation1}.

Prompt augmentation had the ability to make aspects of the given text prompt more prevalent. In the top example of Figure~\ref{fig:ablation1}, the augmented information added to the prompts all contained clues to guide Stable Diffusion to create a more modern looking image of a street in the different cultures.

\subsection{Quantitative Results}

Following our experimental setup detailed in Section~\ref{sec:experiments}, we present our results from our survey in Table~\ref{tab:quant_results}. We included five countries in our survey based on the number of evaluators we could recruit: China, Mexico, Korea, the United States, and Nigeria. Overall, 72 people participated in the surveys with 2,244 image comparisons made.

\noindent
\textbf{Fine-Tuned Stable Diffusion}
The top row of Table~\ref{tab:quant_results} shows our survey results for our fine-tuned Stable Diffusion model versus the baseline. This approach greatly improved all three metrics overall; text-image alignment and cultural alignment were increased and offensiveness greatly decreased.

\noindent
\textbf{Prompt Augmentation}
Prompt augmentation greatly improved cultural alignment over the baseline; however, the images were found to be equally offensiveness to the baseline. 

\noindent
\textbf{Combined}
Our combined approach is better than the baseline in all three metrics. It finds a middle-ground in performance versus its individual parts for text-image alignment and offensiveness; however, overall it performs best in cultural-alignment.

\section{Discussion}

In our ablation survey results shown in Table~\ref{tab:quant_results}, we found that  prompt augmentation decreased text-image alignment. Prompt augmentation makes alterations to the given prompt as seen in Figure~\ref{fig:ablation1} which we believe contributes to the text-image alignment performance diminishment. The augmentation is designed to add culturally relevant information to the prompt, but this added information does not necessarily overlap with the content that Stable Diffusion is capable of generating. For example, in Figure~\ref{fig:ablation1}, the middle column of the bottom example adds the Korean instrument ``gayageum'' to the text prompt with augmentation.  The generated images failed to accurately depict a gayageum which hurts the text-image alignment when compared to to the baseline result which did not include that instrument.


Combining prompt augmenting with our fine-tuned Stable Diffusion model performed better in cultural alignment than the two techniques separately. We hypothesize that the fine-tuned Stable Diffusion learnt concepts that were relevant to the information that the prompt augmentation added, therefore, expanding the content that Stable Diffusion can generate as in the Chinese Pipa instrument example.
Further experiments are needed to thoroughly explain this result.

\subsection{Limitations}

Our two proposed approaches for adapting an existing text-to-image model to be more culturally relevant were successful versus a baseline and Japanese Stable Diffusion; however, we believe that there is more progress to be made. In particular, having guarantees on avoiding offensive generated images is paramount to having trust in using these systems in the wild.

Our representation of culture is nation based in this paper. Most nations have multiple cultures, and culture can exist outside of geographic borders. 

The current version of CCUB dataset was adequate for our experimental purposes. To further improve the quality of CCUB, the number of cultural experts would need to be increased to cross-validate the data. 

In future work, we look to adapt our method to allow for a more personal sense of culture where a user of the system can upload their own visual and textual data according to their own culture to tailor the text-to-image system to themselves.

\section{Conclusion}

We introduce the task of culturally-aware image synthesis and present its importance based on literature results demonstrating the importance of accurate representation in media. As artificial intelligence improves and is used to produce more visual content, cultural representation within this generated media is increasingly important. 

To support culturally-aware text-to-image synthesis we collected the Cross-Cultural Understanding Benchmark (CCUB) dataset consisting of images and captions for 8 different country's cultures collected by people who are a part of each of those cultures. We present two techniques for using this data to alter an existing text-to-image model to be less offensive and more culturally relevant. Our two techniques alone and combined outperform a baseline of specifying the culture in the text prompt. 

Towards equitable representation in text-to-image synthesis, we plan to continue building CCUB to include more cultures in the future. 

\appendix

\section*{Ethical Statement}

While our approach to culturally-aware text-to-image synthesis shows promise in mitigating offensiveness and accurately representing culture, it can still make mistakes and show harmful stereotypes. Additionally, the model still carries over other forms of bias such as gender and racial biases that the large text-to-image models are known to have. For these reasons, we would not recommend this work to be used for cases where people could be harmed by the generated images.

\section*{Acknowledgments}

This work is in part supported by NSF IIS-2112633. Youeun Shin, Youngsik Yun, and Jihie Kim were supported by the MSIT (Ministry of Science, ICT), Korea, under the High-Potential Individuals Global Training Program (RS-2022-00155054) supervised by the IITP (Institute for Information \& Communications Technology Planning \& Evaluation) (50\%).

We would like to thank Ingrid Navarro, Nariaki Kitamura, Sindre Stoe Hobber, Huy Quyen Ngo, Yu Chen, and many others for their help with our CCUB dataset.

\bibliographystyle{named}
\bibliography{ijcai23}

\clearpage
\section*{Appendix}
\subsection*{Prompt Augmenting}

\begin{figure}[h]
\centering
\includegraphics[width=0.95\columnwidth]{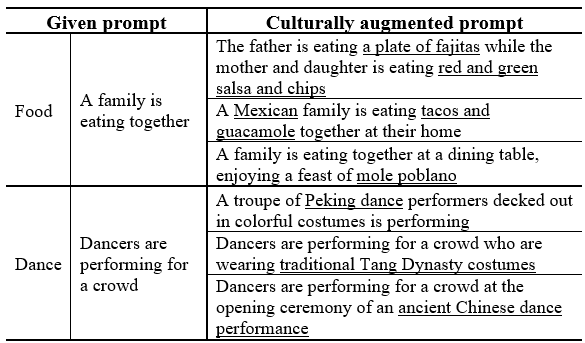}
\caption{Variety of augmented prompts for Mexican food and Chinese dance. The cultural-specific words are underlined. The prompt used for fine-tuned Stable Diffusion is a given prompt plus a culturally augmented prompt appended by its culture name.}
\label{fig::various_aug}
\end{figure}

In this section, we describe the flexibility of our prompt engineering approach with several samples of augmentation outputs. 
We demonstrate that our approach can augment a given prompt in various ways. Some examples of food and dance in Mexican and Chinese culture, respectively, are listed in 
Figure~\ref{fig::various_aug}.
The augmenting process adds semantic description in detail as well as the cultural context of the corresponding culture. It is shown that our approach can supplement the information that is not in our training data, benefiting from the immense knowledge of GPT-3. 

\begin{figure}[h]
\centering
\includegraphics[width=0.95\columnwidth]{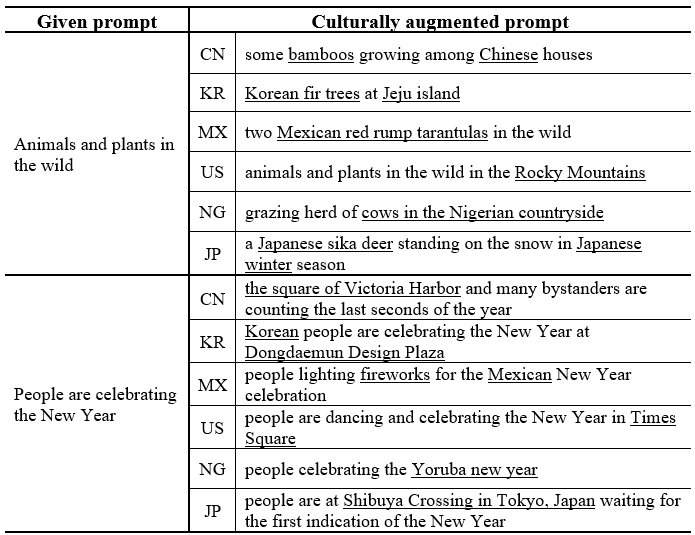}
\caption{Examples of augmented outputs beyond our cultural categories in the order of China, Korea, Mexico, the United States, Nigeria, and Japan. The cultural-specific words are underlined.}
\label{fig::augsample}
\end{figure}

Our fine-tuned large language model is also capable of generating relevant cultural context beyond the defined nine cultural categories. As shown in Figure~\ref{fig::augsample}, we try two types of new cultural elements, nature in the wild and a new year celebration, out of the nine categories. Prompt augmented results for all six cultures we address in our paper consist of explicit cultural context, such as well-known cultural landmarks and species originating from those cultures.

\subsection*{Additional Culturally-aware Text-to-image Synthesis Results}

\begin{figure*}[h!]
    \centering
    \includegraphics[width=1\textwidth]{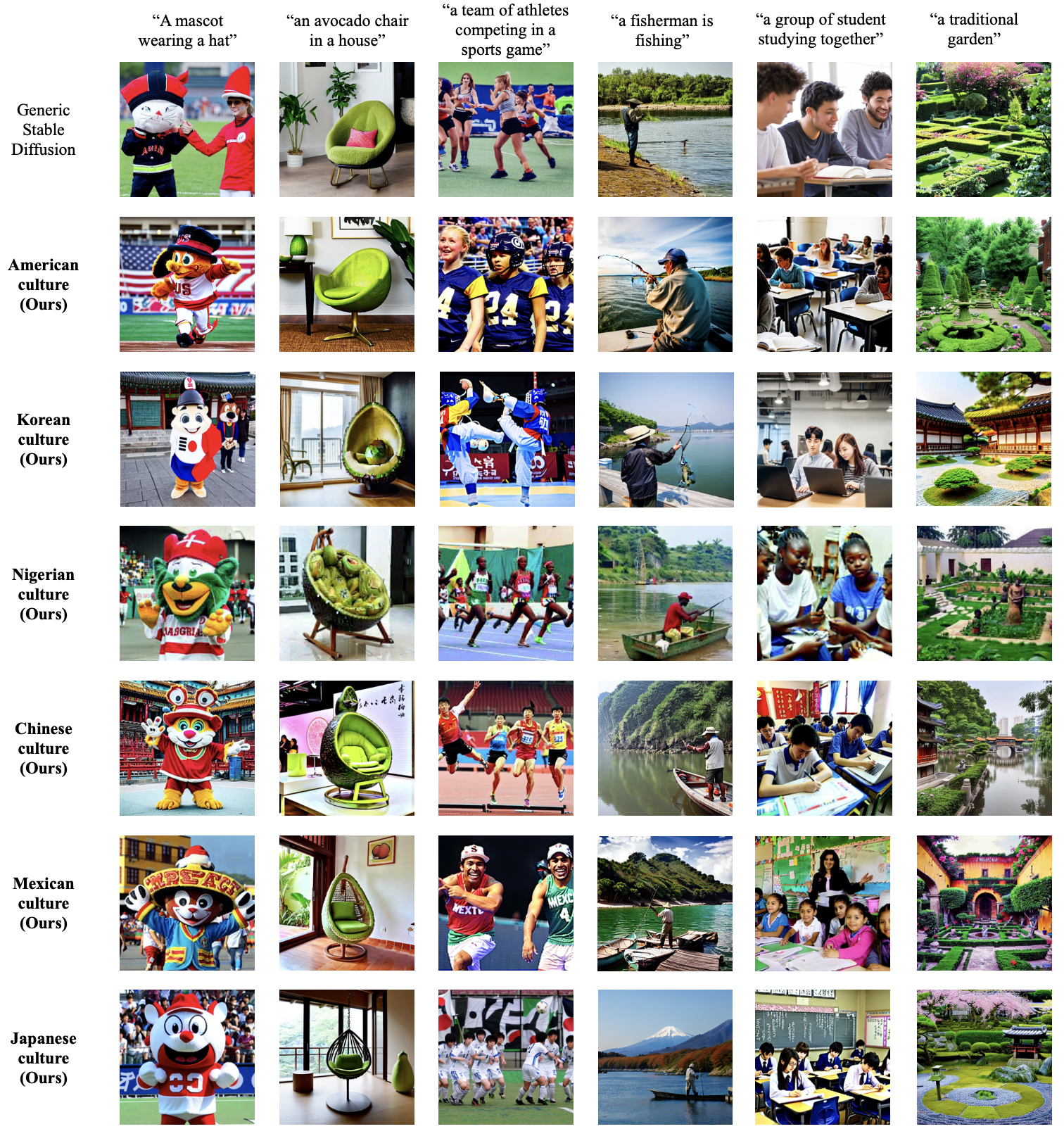}
    \caption{More results of our approach to achieving culturally-aware text-to-image synthesis for six cultures. The first row is the text prompts used, the first column is the culture condition.
    }
    \label{fig:front_apx}
\end{figure*}

In Figure~\ref{fig:front_apx}, more results of our approach to achieving culturally-aware images given a text prompt appended by a culture name are shown. We prove the generalizability of our model conditioned on diverse text prompts and demonstrate that our approach is able to generate culturally aligned images within and beyond the nine categories we defined for a culture.

\subsection*{Cultural Bias in Stable Diffusion 2}

\begin{figure}[h]
\centering
\includegraphics[width=1\columnwidth]{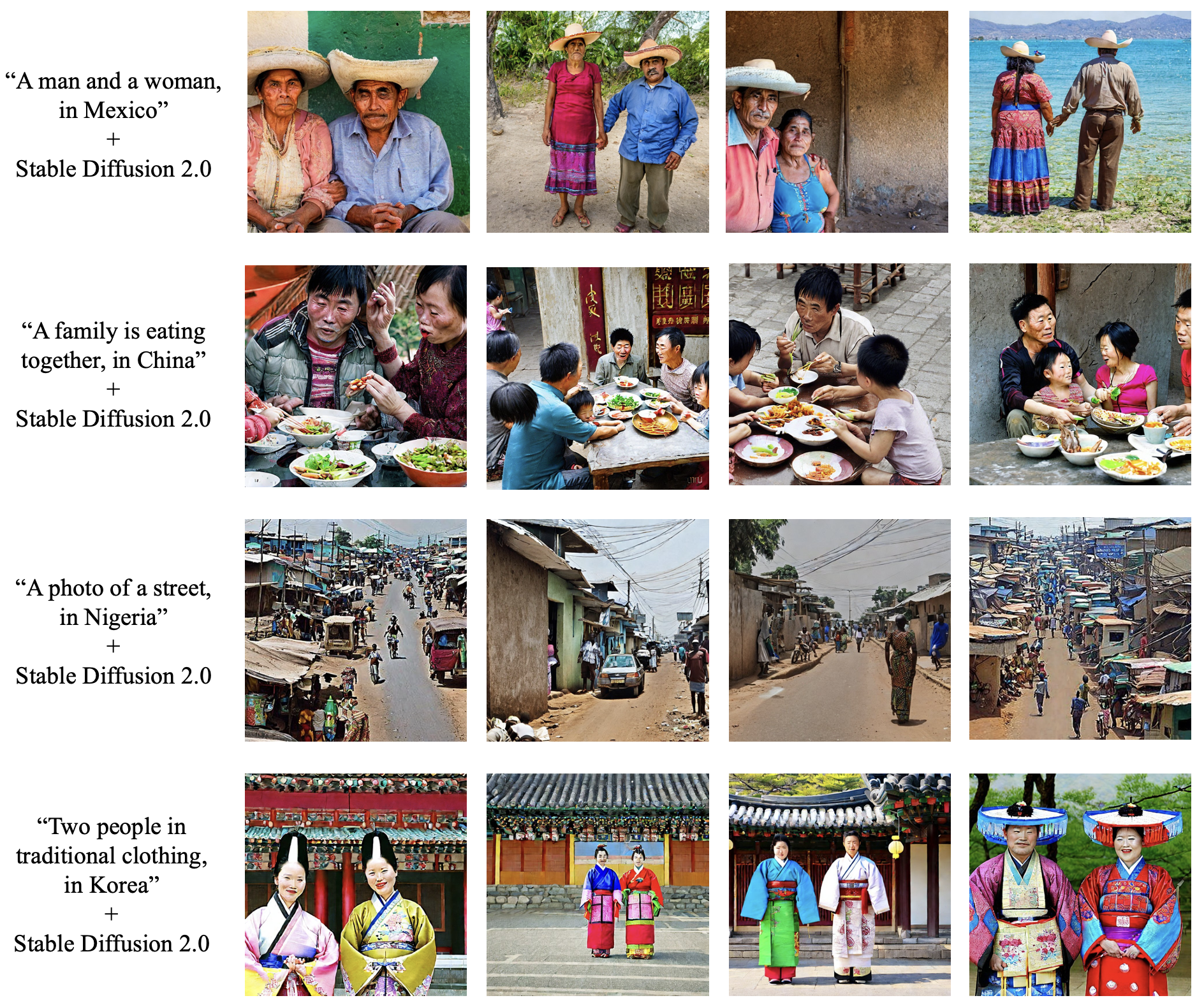} 
\caption{Random samples generated by generic Stable Diffusion version 2.0. The first column shows the prompts.}
\label{fig::sd2}
\end{figure}

In our approach, we fine-tune Stable Diffusion~\cite{rombach2021stableDiffusionOriginal} version 1.4. While Stable Diffusion 2.0 has been released recently, intrinsic cultural biases yet commonly exist. In Figure~\ref{fig::sd2}, we show randomly generated images based on four prompts. We claim that Stable Diffusion 2.0: 1) generates images containing stereotypes for different cultures: in the first and second row in Figure~\ref{fig::sd2},  the facial expressions for Chinese and Mexican culture are stereotypical and offensive; 2) generates discriminative items: given the prompt ``A photo of a street, in Nigeria", Stable Diffusion 2.0 is also prone to generate old-looking and discriminative images for Nigerian culture; 3) contains cultural misrepresentations:  in the images with the prompt ``Two people in traditional clothing, in Korea", the traditional clothing is more in a Japanese style instead of Korean style.

\end{document}